\documentclass[letterpaper]{article}
\usepackage[preprint]{aaai2027}
\usepackage[hyphens]{url}
\usepackage{graphicx}
\usepackage{natbib}
\usepackage{caption}
\usepackage{amsmath}
\usepackage{booktabs}
\usepackage{multirow}
\usepackage{algorithm}
\usepackage{algpseudocode}

\title{TROVE: Adaptive Agent Skill Orchestration via Trace-Grounded Route Validation and Editing}
\author{Tianxing Wang, Mingming Zhao, Shuai Huang, Huiyang Xu,\newline
Chaoyue Niu, Shengzhong Liu, Fan Wu}
\affiliations{}

\begin{document}

\maketitle

\begin{abstract}
Agents tend to optimize, select, or constrain execution structures before decisive runtime outcomes are observed. However, such pre-execution commitment creates an orchestration bottleneck: when intermediate evidence invalidates the pending continuation, agents must either execute stale steps or replan broadly, compounding errors, wasting computation, and discarding progress. We thus propose Trace-grounded Route Orchestration via Validation and Editing (TROVE), which revises only what runtime evidence invalidates. Offline, TROVE distills evaluated workflow-search traces into atomic and composite skills and an outcome-conditioned transition graph, preserving stable fragments while exposing outcome-dependent decisions. Online, it treats a planned route as provisional: after committing one top-level skill, the controller retains a valid continuation, inserts a trace-supported local response, or replaces only the invalid suffix. Evaluations across code-generation, question-answering, and math reasoning benchmarks with different LLM backbones show that TROVE delivers a stronger quality-efficiency trade-off than existing baselines of dataset-level optimization, query-level architecture selection, and graph-constrained scheduling. Quality gains are largest when outcomes change the appropriate continuation, whereas early termination yields substantial efficiency gains on near-saturated tasks. Ablations further show that composite skills capture most offline benefits, insertion enables local correction, and suffix replacement primarily improves efficiency. These findings establish selective route editing as a general principle for adaptive agent orchestration.
\end{abstract}

\section{Introduction}
\label{sec:introduction}

Agents increasingly solve complex tasks by orchestrating multiple reasoning, generation, verification, and tool-use steps. The task performance depends not only on the capability of each component, but also on when and in what order those components are invoked. A strong solver may need a critic only when its answer is uncertain; a generated program may require repair only after a failed test; and a planned verification step becomes redundant once an earlier skill has already produced a valid final answer. Agent workflow orchestration is therefore not merely a mechanism for composing model calls. It determines whether intermediate evidence is converted into useful computation or ignored by a stale execution path.

Existing work has substantially improved how agent workflows are constructed, but places most structural decisions before the outcomes that matter become observable. For example, AFlow selects a dataset-level workflow~\cite{zhang2025aflow}; MaAS chooses a query-specific architecture before execution~\cite{zhang2025maas}; and LAS adapts only within a predefined graph~\cite{xiang2026las}. However, as execution unfolds, a queued continuation can become stale: review may demand revision, success may render verification redundant, or tool failure may invalidate downstream steps. Executing that continuation compounds errors and wastes computation.
\begin{figure}[t]
  \centering
  \includegraphics[width=\columnwidth]{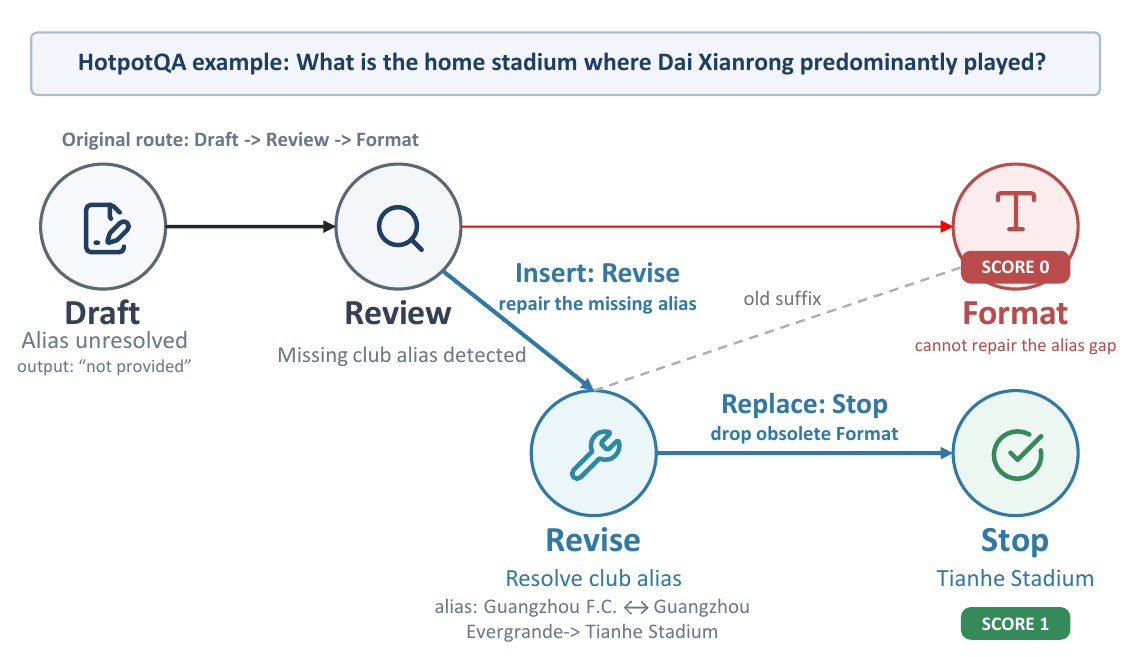}
  \caption{Motivating HotpotQA example. After \texttt{review} (2) detects missing evidence in \texttt{draft} (1), AFlow still executes its fixed \texttt{format} suffix and scores 0. Our TROVE instead applies \textsc{Insert} to add \texttt{revise} (3), followed by \textsc{Replace} to substitute \texttt{stop} (4) for the obsolete suffix, preserving the useful prefix and raising the score to 1.}
  \label{fig:motivation-case}
\end{figure}

Such runtime mismatch motivates the new requirement of adaptive route validation and editing: after each execution boundary, validate the pending route and edit it at the smallest sufficient scope. Unlike restarting or regenerating the entire future, this strategy preserves the executed prefix and artifacts, retains valid steps, repairs local mismatches, and replans only an invalid suffix. Figure~\ref{fig:motivation-case} illustrates the idea on HotpotQA: \texttt{review} exposes missing evidence, but AFlow still executes the queued \texttt{format} step and fails. We instead insert \texttt{revise} and replace the obsolete suffix with \texttt{stop}, preserving the useful \texttt{draft}-\texttt{review} prefix while changing only what the new evidence invalidates.

Building on this insight, we propose Trace-grounded Route Orchestration via Validation and Editing, \textbf{TROVE} for short. Offline, TROVE distills evaluated workflow search traces into a skill registry and a transition graph: stable fragments become executable composite skills, whereas outcome-dependent continuations remain explicit. Online, a planner proposes a short route, but the controller commits to only its first top-level skill. After observing outcome, TROVE retains the planned successor if it remains valid, inserts a local response if the downstream intent remains sound, or invokes an LLM to replace only the invalid suffix. TROVE thus separates planning from commitment: its proposal horizon spans multiple skills, while its commitment horizon remains one action.

We summarize the contributions of this work as follows:
\begin{itemize}
    \item We identify a fundamental agent orchestration challenge: adapting a workflow when runtime outcomes invalidate what it planned to do next.

    \item We propose TROVE, which distills evaluated search traces into executable composite skills and an outcome-conditioned transition graph, then combines one-step commitment with different route revisions to
    preserve useful structure and artifacts.

    \item We evaluate TROVE on 6 code, question-answering, and
    mathematics benchmarks with 3 LLMs. Across 18 backbone-benchmark settings, TROVE achieves the best or tied-best task score in 15 and the shortest online time in 16. Relative to
    AFlow, TROVE improves task score in 16 settings (by up to 31.37\%) and reduces time in 16 (by up to 86.7\%).
\end{itemize}

\section{Related Work}

\subsection{Workflow Construction Before Execution}

Automated workflow construction spans {\em operation-}, {\em workflow-},
and {\em query-level} structure. At the operation level, GPTSwarm
represents agent systems as computational graphs and optimizes
node prompts and connectivity through task feedback
\cite{zhuge2024gptswarm}. At the workflow and system level,
AFlow searches executable code workflows, while ADAS programs
complete agent systems from an archive of prior designs
\cite{zhang2025aflow,hu2025adas}. AFlow is representative of
this level: it uses execution feedback to optimize a
dataset-level workflow that is subsequently reused at test time.
Query-level methods instead construct or select an architecture
for each instance, including MaAS, DyLAN, and FlowReasoner
\cite{zhang2025maas,liu2023dylan,gao2025flowreasoner}.
MaAS, for example, samples a customized multi-agent
architecture from an optimized agentic supernet before
execution. Across these levels, the main structure is determined
before instance-specific intermediate outcomes are observed.
TROVE instead treats a proposed skill route as provisional and
revises only its unexecuted continuation online.

\subsection{Skill Abstraction and Experience Reuse}

Skill-oriented methods differ in how reusable capability is
obtained and represented. Classical temporal abstraction and
SayCan assume predefined executable policies or skills, with
SayCan selecting actions according to both linguistic relevance
and environmental feasibility
\cite{sutton1999options,ahn2022saycan}. Code as Policies moves
from selection to synthesis by composing perception and control
APIs into executable programs with functions, conditions, and
loops \cite{liang2023codeaspolicies}. Experience-based methods
derive reuse from previous interactions: ExpeL extracts textual
insights from trajectories, whereas Voyager stores acquired
behaviors as an executable code-skill library
\cite{zhao2024expel,wang2023voyager}. Recent systems further
organize and compose large skill collections through capability
trees, executable DAGs, or evolving skill-relation graphs
\cite{li2026agentskillos,xia2026grasp,li2026skillgraph}.
TROVE similarly operationalizes experience, but derives its
units specifically from evaluated workflow-search traces:
existing operators become atomic skills, stable recurring
fragments become composite skills, and outcome-dependent
continuations remain explicit in the transition graph.

\subsection{Execution-Time Planning and Route Adaptation}

Execution-time controllers differ in how much future structure
they preserve after feedback. ReAct generates only the next
action after each observation
\cite{yao2023react}. Plan-level methods such as AdaPlanner and
DoReMi retain a high-level plan and refine or recover it when
execution feedback violates its assumptions
\cite{sun2023adaplanner,guo2023doremi}. State- and
workflow-level controllers instead restrict adaptation to
structured transition spaces: StateFlow routes execution through
a state machine, LAS dynamically selects paths within a workflow
DAG, and FlowSwitch decides when to continue or switch
workflows according to dialogue state
\cite{wu2024stateflow,xiang2026las,chang2026flowswitch}.
LAS is the closest representative, using intermediate outputs to
selectively exit, verify, repair, or reroute within an existing
workflow. TROVE instead edits the unresolved suffix of a
task-specific provisional route: it retains a valid successor,
inserts a local transition-supported response when possible, and
replaces the suffix only when broader replanning is required.
This preserves the executed prefix and still-valid route intent
without restricting adaptation to a predefined DAG.

\begin{figure*}[t]
  \centering
  \includegraphics[width=\textwidth]{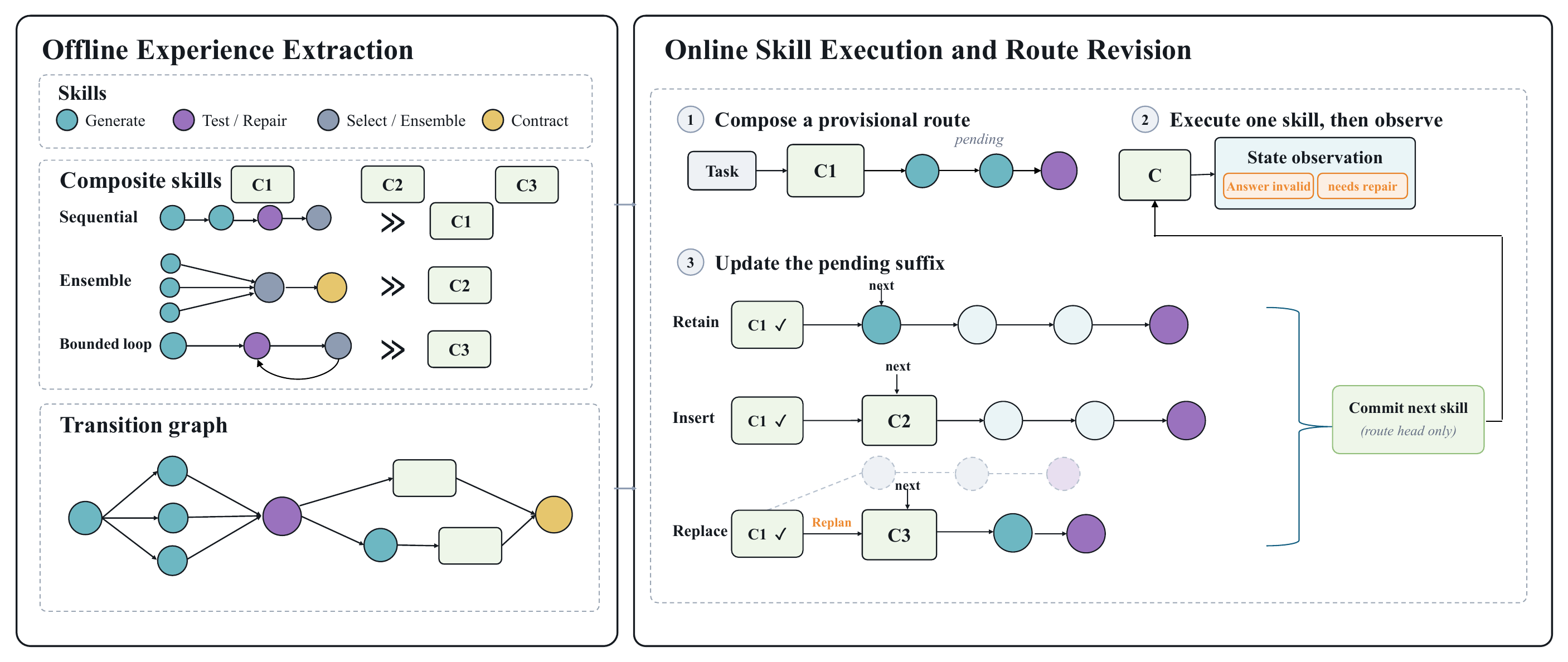}
  \caption{Offline experience extraction and online route revision in TROVE.
  Offline, standardized atomic skills are organized into stable composite
  skills and a transition graph (left).  Online, a planner proposes a
  provisional route, but the controller executes only its head and observes
  the resulting state.  The pending suffix is then retained, extended by
  inserting a locally supported skill, or replaced through replanning.  Only
  the selected next skill is committed before the execution--observation loop
  repeats (right).}
  \label{fig:trove-overview}
\end{figure*}

\section{Design of TROVE}
\label{sec:design}

\subsection{Problem Formulation and Overview}

\paragraph{Problem Formulation.}
Let $\mathcal{S}$ be a registry of executable skills and
$R=(s_1,\ldots,s_k)$ a skill route proposed before its
intermediate outcomes are observed. After executing $s_t$,
the latest outcome may invalidate the pending continuation:
the planned successor may lack a required artifact, a queued
repair may become unnecessary, the route may be exhausted
before the task is complete, or repeated actions may cease to
make progress. We refer to this mismatch between the
proposed continuation and the observed execution state as
\emph{continuation invalidation}.

Existing systems commonly address this problem at two
extremes. A coarse-grained strategy abandons the current
workflow, selects another complete workflow, and re-executes
the task. Although this may replace an unsuitable global
structure, it also discards useful progress and repeats
completed model calls, tool executions, and intermediate
artifacts. At the other extreme, step-wise LLM replanning
selects a new action or regenerates the remaining route after
every node. This provides greater reactivity, but repeatedly
incurs routing latency and token cost, introduces decision
variance, and may overwrite portions of the original route
that remain valid.

TROVE occupies the middle ground between these strategies.
It treats a planner-generated route as provisional intent and
commits to only one top-level skill at a time. After observing
the outcome of the committed skill, TROVE retains the
planned continuation when it remains valid, inserts a bounded
local response when the original intent can still be preserved,
and replaces only the unexecuted suffix when broader
replanning is necessary.

\paragraph{Method Overview.}
TROVE separates offline experience construction from online
route adaptation. Offline, AFlow operators are standardized
as atomic skills, and evaluated one-node search traces are
organized into two complementary forms of reusable
experience: stable local fragments are encapsulated as
composite skills, while outcome-dependent continuations are
retained in a skill transition graph.

Online, a planner uses the resulting skill registry and
transition graph to propose a short route. The controller
executes only its first skill and, after observing the boundary
outcome, decides whether to terminate, retain the queued
successor, insert a local response, or invoke an LLM to
replace the unexecuted suffix. Thus, the proposal horizon
may cover multiple skills, while the commitment horizon
remains one top-level action.
\subsection{Offline Experience Extraction}

TROVE starts from the operators used by AFlow. Since each
operator is an atomic execution node in a workflow, we wrap
it as an atomic skill with a standardized interface describing
its required inputs, produced artifacts, execution status, and
repeat constraints. These skills form the initial registry
$\mathcal{S}_A$ and the primitive action space of offline
workflow search.

AFlow searches workflows on the training partition. In each
iteration, it modifies one workflow node by adding or
substituting an operator and evaluates the resulting workflow.
Let $W_i$ and $W_{i+1}$ denote two consecutive candidates.
The corresponding changes are
\begin{equation}
\begin{aligned}
\Delta_i^{\mathrm{score}}
    &= \operatorname{Score}(W_{i+1})
       - \operatorname{Score}(W_i), \\
\Delta_i^{\mathrm{time}}
    &= \operatorname{Time}(W_{i+1})
       - \operatorname{Time}(W_i).
\end{aligned}
\end{equation}
Together with the executed skill sequence and intermediate
outcomes, these paired evaluations provide localized
comparative evidence about how a structural modification
affects task quality and execution cost. TROVE therefore
reuses the search trajectory rather than retaining only the
final workflow selected by AFlow.

\paragraph{Atomic and Composite Skill Abstraction.}
Atomic skills remain the primitive execution units. Using
the operator sequences and one-node modifications recorded
during AFlow search, TROVE identifies recurring local
fragments in effective trajectories as composite candidates.
A candidate is promoted to a composite skill when its
internal ordering and data dependencies remain stable, it
implements a coherent subgoal, exposes a consistent
input--output contract, and can execute without consulting
the pending outer route. If an intermediate outcome may
change a later top-level decision, the fragment remains
decomposed so that the corresponding boundary is preserved.

Accepted fragments form the composite set
$\mathcal{S}_C$, yielding
\begin{equation}
    \mathcal{S}=\mathcal{S}_A\cup\mathcal{S}_C.
\end{equation}
Atomic and composite skills expose the same top-level
interface. Once invoked, a composite executes its child
skills internally and returns one aggregate boundary outcome
to the outer controller. Their occurrences are collapsed into composite nodes in the
search traces, aligning offline experience with the
top-level actions used by the online controller.

Composite abstraction serves a specific role in TROVE: it hides
stable local decisions behind one execution boundary while
leaving outcome-dependent continuations exposed to the online
controller.

\paragraph{Skill Transition Graph.}
The same search traces also reveal relations that should not
be hidden inside a composite. When the appropriate
continuation depends on an observed outcome, TROVE retains
the relation explicitly in a skill transition graph. After
accepted fragments are mapped to composite nodes, a
transition is represented as
\begin{equation}
    \tau=(s,e,c,g), \qquad s,g\in\mathcal{S},
\end{equation}
where $s$ is the completed top-level skill, $e$ is the
observed boundary event, $c$ summarizes the observable
boundary features shared by the supporting trace fragments,
and $g$ is one supported immediate response skill.

An edge is derived from an evaluated local workflow
modification rather than from skill adjacency alone. The
corresponding score and time changes serve as offline
comparative evidence for retaining the proposed response;
they are not used as routing inputs during online execution.

At deployment, the graph serves two bounded roles. Before
execution, the planner uses nominal continuation hints
supported by task-visible context. After a skill completes,
graph lookup additionally conditions on the observed
boundary outcome and provides a next-skill candidate when
the pending route becomes invalid. The graph does not
construct a complete workflow or directly edit the remaining
suffix. Whether the retrieved candidate should be inserted
locally or used as a prior for LLM-based suffix replacement
is decided by the online controller.

\subsection{Online Skill Execution and Route Revision}

\paragraph{Provisional Route and One-Step Execution.}
At deployment, a planner uses the task $x$, the
offline-constructed skill registry $\mathcal{S}$, and the
skill transition graph $\mathcal{G}$ to propose a short route
\begin{equation}
    R_0=\operatorname{Plan}(x,\mathcal{S},\mathcal{G})
       =(s_1,\ldots,s_k).
\end{equation}
The graph provides bounded continuation hints rather than a
complete path. Accordingly, the route is treated as
provisional intent: TROVE commits to only its first
top-level skill and keeps the remaining skills as an
unexecuted suffix.

At iteration $t$, let $R_t = (s_t) \oplus Q_t$, where $s_t$ is the
committed skill, $Q_t$ is the provisional suffix, and $H_t$
is the preceding execution history. TROVE executes $s_t$,
observes its boundary outcome $o_t$, and updates the history
as $H_{t+1}=H_t\oplus(s_t,o_t)$. The outcome summarizes
the execution status, available artifacts, and signals
relevant to the next routing decision.

\paragraph{Observation-Guided Route Update.}
If the boundary outcome indicates that the task is complete,
TROVE returns the current result and discards the pending
suffix. Otherwise, it examines the queued successor.
\emph{Retain} is used when that successor remains executable
under the latest outcome: the controller preserves the
original suffix and continues without another LLM routing
call.

If the successor is absent or no longer applicable, TROVE
queries the transition graph using the completed skill and
the observed boundary state. The graph returns one
trace-supported response candidate when available and
abstains otherwise. We denote the lookup result by
$g_t \in S \cup \{\bot\}$,
where $g_t=\bot$ indicates graph abstention. The controller then revises the route through one
of two operations.

\emph{Insert} is used when $g_t$ addresses the immediate
mismatch---for example, through repair, testing, or
verification---without changing the downstream objective
encoded by the pending suffix. TROVE places the response
before that suffix:
\begin{equation}
    R_{t+1}=(g_t)\oplus Q_t.
\end{equation}
After $g_t$ completes, the original successor is checked
again rather than executed automatically. When $Q_t$ is
empty, the same operation acts as a bounded route extension.

\emph{Replace} is used when $g_t=\bot$, when the graph
evidence is ambiguous, or when the observed state requires
a broader change to the continuation. It is also used when
a graph-supported response indicates a new direction that
cannot preserve the downstream objective of the pending
suffix. An LLM replanner receives the task, the latest
outcome, the updated execution history, the pending suffix,
the available skills, and $g_t$ as a routing prior when it is
available. It then generates a bounded replacement suffix:
\begin{equation}
    \widehat{Q}_t
    =\operatorname{Replan}
      (x,o_t,H_{t+1},Q_t,\mathcal{S};g_t),
    \qquad
    R_{t+1}=\widehat{Q}_t.
\end{equation}
When $g_t=\bot$, the replanner proceeds without a graph
recommendation. Unlike workflow reselection and
re-execution, replacement preserves the completed prefix
and its artifacts and regenerates only the unexecuted
portion.

\paragraph{Iterative Execution.}
TROVE repeatedly commits to the first skill of the updated
route and applies the same outcome-guided update after each
boundary. Execution terminates when a final result is
produced or when the bounded step, repetition, or
no-progress constraints are reached. If a budget or no-progress bound is reached before a valid
final result is produced, the execution is marked incomplete
and scored according to the benchmark evaluator.

\section{Evaluation}
\label{sec:evaluation}

\begin{table*}[t]
  \centering
  \caption{Combined task quality and online elapsed time on the held-out test
  partitions. Score is reported in percent and time in minutes. HumanEval and
  MBPP use pass@1, DROP and HotpotQA use mean token F1, and MATH and GSM8K use
  accuracy. Within each backbone and benchmark, \textbf{bold} and
  \underline{underlining} independently mark the best and second-best value
  for each metric. ``--'' indicates that LAS does not publicly release a
  trained gate for that non-code benchmark.}
  \label{tab:main-results}
  \resizebox{\linewidth}{!}{
  \begin{tabular}{@{}ll*{6}{cc}@{}}
    \toprule
    \multirow{3}{*}{Backbone} & \multirow{3}{*}{Method} &
      \multicolumn{4}{c}{Code} &
      \multicolumn{4}{c}{QA} &
      \multicolumn{4}{c}{Math} \\
    \cmidrule(lr){3-6}\cmidrule(lr){7-10}\cmidrule(lr){11-14}
      & &
      \multicolumn{2}{c}{HumanEval} &
      \multicolumn{2}{c}{MBPP} &
      \multicolumn{2}{c}{DROP} &
      \multicolumn{2}{c}{HotpotQA} &
      \multicolumn{2}{c}{MATH} &
      \multicolumn{2}{c}{GSM8K} \\
    \cmidrule(lr){3-4}\cmidrule(lr){5-6}\cmidrule(lr){7-8}
    \cmidrule(lr){9-10}\cmidrule(lr){11-12}\cmidrule(lr){13-14}
      & & Score $\uparrow$ & Time $\downarrow$
      & Score $\uparrow$ & Time $\downarrow$
      & Score $\uparrow$ & Time $\downarrow$
      & Score $\uparrow$ & Time $\downarrow$
      & Score $\uparrow$ & Time $\downarrow$
      & Score $\uparrow$ & Time $\downarrow$ \\
    \midrule
    \multirow{4}{*}{\textit{DeepSeek-V4-Flash}}
      & AFlow~\cite{zhang2025aflow} & 93.89 & 21.75 & 91.79 & 43.10 &
        \underline{88.50} & \underline{15.56} &
        \underline{78.58} & \underline{25.41} &
        \underline{74.07} & \underline{68.88} &
        96.02 & \underline{57.73} \\
      & MaAS~\cite{zhang2025maas}  & 87.02 & \underline{13.52} & 81.82 & 35.13 &
        81.64 & 22.98 & 77.37 & 26.60 & 68.11 & 131.83 &
        \underline{96.68} & 108.23 \\
      & LAS~\cite{xiang2026las}   & \underline{94.66} & 13.81 &
        \underline{92.08} & \underline{22.02} &
        -- & -- & -- & -- & -- & -- & -- & -- \\
      & {\bf Our TROVE} & \textbf{97.71} & \textbf{8.10} &
        \textbf{93.55} & \textbf{12.63} &
        \textbf{90.39} & \textbf{14.89} &
        \textbf{80.16} & \textbf{23.97} &
        \textbf{80.86} & \textbf{26.08} &
        \textbf{97.16} & \textbf{9.63} \\
    \midrule
    \multirow{4}{*}{\textit{GPT-4o-mini}}
      & AFlow~\cite{zhang2025aflow} & \underline{93.13} & 5.18 & 75.95 & 9.42 &
        \underline{65.61} & \textbf{7.83} &
        \underline{68.46} & \underline{11.26} &
        \textbf{52.26} & \underline{36.76} &
        \underline{92.23} & \underline{37.79} \\
      & MaAS~\cite{zhang2025maas}  & \textbf{95.42} & 8.19 & 74.49 & 20.68 &
        34.55 & 13.13 & 57.90 & 11.73 & 51.44 & 122.04 &
        \textbf{92.80} & 149.02 \\
      & LAS~\cite{xiang2026las}   & 84.73 & \underline{2.94} &
        \textbf{80.65} & \underline{6.01} &
        -- & -- & -- & -- & -- & -- & -- & -- \\
      & {\bf Our TROVE} & 89.31 & \textbf{1.76} &
        \underline{78.59} & \textbf{2.17} &
        \textbf{80.82} & \underline{8.06} &
        \textbf{70.05} & \textbf{10.93} &
        \underline{52.06} & \textbf{8.01} &
        \textbf{92.80} & \textbf{8.04} \\
    \midrule
    \multirow{4}{*}{\textit{Qwen3-8B}}
      & AFlow~\cite{zhang2025aflow} & \underline{90.08} & 60.53 & 51.91 & 116.58 &
        \underline{85.46} & \underline{42.48} &
        \underline{75.91} & \textbf{44.05} & 8.44 & 273.35 &
        \underline{87.04} & \underline{228.38} \\
      & MaAS~\cite{zhang2025maas}  & 65.65 & \underline{43.11} & 62.76 & 96.84 &
        63.04 & 67.92 & 66.17 & 57.82 &
        \underline{12.55} & \underline{235.70} & 76.21 & 346.33 \\
      & LAS~\cite{xiang2026las}   & 89.31 & 44.08 &
        \underline{76.25} & \underline{66.75} &
        -- & -- & -- & -- & -- & -- & -- & -- \\
      & {\bf Our TROVE} & \textbf{92.37} & \textbf{24.36} &
        \textbf{83.28} & \textbf{41.84} &
        \textbf{85.85} & \textbf{37.23} &
        \textbf{76.03} & \underline{54.27} &
        \textbf{36.42} & \textbf{36.41} &
        \textbf{92.80} & \textbf{44.18} \\
    \bottomrule
  \end{tabular}
  }
\end{table*}

\subsection{Experimental Setup}

\paragraph{Benchmarks and Data Partitions.}
We evaluate on six benchmarks spanning code generation, mathematical
reasoning, and question answering: HumanEval \citep{chen2021humaneval}, MBPP
\citep{austin2021mbpp}, MATH \citep{hendrycks2021math}, GSM8K
\citep{cobbe2021gsm8k}, DROP \citep{dua2019drop}, and HotpotQA
\citep{yang2018hotpotqa}. Each benchmark is divided approximately 1:4 into a
training partition and a held-out test partition. The resulting
training/test sizes are 33/131 for HumanEval, 86/341 for MBPP, 119/486 for
MATH, 264/1,055 for GSM8K, and 200/800 for both DROP and HotpotQA. Workflow
search, controller configuration, trace collection, composite-skill
construction, and transition-memory construction use only the training
partition. All test-time structures are frozen before the corresponding test
partition is evaluated.

\paragraph{Baselines.}
We compare TROVE with three representative workflow
orchestration methods operating at different adaptation
granularities. AFlow~\cite{zhang2025aflow} searches a
dataset-level workflow over operator nodes on the training
set and executes the selected workflow unchanged at test
time. It represents offline workflow optimization and
evaluates whether outcome-aware route revision improves
over replaying a strong fixed workflow. MaAS~\cite{zhang2025maas}
samples a query-specific multi-agent
architecture from an optimized agentic supernet before
execution. It allows the workflow structure to vary across
queries, but does not adjust it according to intermediate
outcomes, providing a comparison with pre-execution
adaptation. LAS~\cite{xiang2026las} dynamically selects the
next node or path within a workflow DAG using intermediate
outputs. As the closest online baseline, it compares dynamic
next-hop scheduling with TROVE's validation and
retain--insert--replace revision of an interruptible skill
route.

Together, these baselines cover dataset-level offline
optimization, query-level pre-execution adaptation, and
node-level online scheduling.
\paragraph{Metrics.}
We report task quality, online execution time, and online token usage.
\emph{Task score} is pass@1
for HumanEval and MBPP, accuracy for MATH and GSM8K, and mean
token-level F1 for DROP and HotpotQA; all values are shown as percentages.
\emph{Time} is the elapsed wall-clock time required to evaluate the complete
test partition under concurrency 10. It includes online planning, skill
execution, route validation, and any runtime replanning, but excludes one-time
offline workflow search, controller training, registry construction, and trace
distillation. \emph{Tokens} are provider-reported input-plus-output totals over
online planner, router, skill, and composite-internal model calls. Scores are
compared only within the same benchmark.

\paragraph{Implementation Details.}
We evaluate AFlow, MaAS, LAS, and TROVE with
DeepSeek-V4-Flash, GPT-4o-mini, and Qwen3-8B whenever
supported by their released implementations. All test runs
use a maximum concurrency of 10 and a 300-second per-sample
timeout.

AFlow deploys workflows selected on the training
partition. MaAS uses its frozen controller,
while LAS uses the released gate checkpoints. All offline
artifacts of TROVE are frozen before test evaluation.

The final TROVE registry contains 10 atomic skills and 41
composite skills, and the deployed transition graph contains
244 executable edges. The initial planner proposes at most
four top-level skills, while online execution is limited to six
top-level skill calls. No test result is used to select a
workflow, controller, or checkpoint.

\subsection{Main Results and Analysis}

\paragraph{Overall Results.}
Table~\ref{tab:main-results} compares all methods under the
same execution backbone. Across the 18 backbone--benchmark
blocks, TROVE obtains the best or tied-best task score in 15
and the shortest online time in 16. Against AFlow, TROVE
improves score in 16 settings, with positive gains of
0.12--31.37 points on the corresponding task metric, and
reduces time in 16 by 2.9--86.7\%. Against MaAS, it improves
or ties score in 17 settings, with positive gains of
0.48--46.27 points, and reduces time in all 18 by
6.1--94.6\%. On the six LAS-supported code settings, it
improves score in five by 1.47--7.03 points and reduces time
in all six by 37.3--63.9\%. The gains reflect different
advantages over each baseline: avoiding stale suffix replay
relative to AFlow, adapting after intermediate outcomes
unavailable to MaAS, and reconstructing the pending route
beyond LAS's predefined DAG.

Across the six DeepSeek-V4-Flash test partitions, TROVE uses
13.45M online tokens versus AFlow's 21.11M, a 36.3\% reduction.
The reduction is driven primarily by MBPP and GSM8K, while
DROP and HotpotQA require more tokens under TROVE.

\paragraph{Differences Across Tasks.}
Performance varies with both task difficulty and where that
difficulty arises. On the harder MATH benchmark, TROVE
gains 6.79 points over AFlow with DeepSeek-V4-Flash while
reducing time by 62.1\%; multi-step reasoning creates more
opportunities for intermediate outcomes to invalidate later
operations. GSM8K is near the performance ceiling, so score
gains are smaller, but early completion still yields large time
savings. HumanEval changes routes more frequently than
MBPP, while code quality remains strongly
backbone-dependent. The moderate gains on DROP and
HotpotQA suggest that their difficulty lies more in evidence
extraction and long-context processing than in unstable
continuations. Overall, adaptation yields larger quality gains
when task difficulty changes what should happen next, but
mainly improves efficiency on high-scoring or evidence-heavy
tasks.

\paragraph{Differences Across Backbones.}
Model capability changes whether adaptation yields quality
gains or mainly efficiency. DeepSeek-V4-Flash shows the most
consistent pattern, with TROVE improving both score and time
on all six tasks, suggesting reliable use of composite
execution and boundary feedback. GPT-4o-mini retains large
time savings but shows mixed scores; strong baseline results
on HumanEval and GSM8K leave less headroom and make the
outcome more sensitive to routing choices. Qwen3-8B gains
most on MBPP and MATH, where lower baseline scores leave
more room for structured execution and selective revision.

\begin{figure}[t]
  \centering
  \includegraphics[width=\columnwidth]{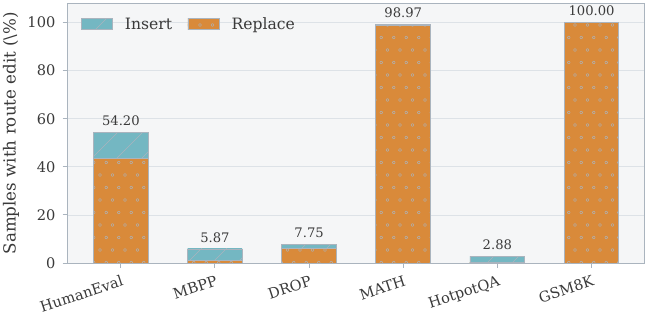}
  \caption{Sample-level incidence of TROVE's online route edits with
  DeepSeek-V4-Flash. Bars stack \textsc{Insert} and \textsc{Replace}; labels
  give the total percentage of samples whose realized route differs from its
  proposal. Terminal pruning counts as replacement, while timeout-incomplete
  paths and identity confirmations are excluded.}
  \label{fig:path-edit-actions}
\end{figure}

\paragraph{Route-Adaptation Frequency.}
Figure~\ref{fig:path-edit-actions} reports how often the
realized route differs from its proposal. HumanEval changes
54.20\% of routes, indicating substantial demand for runtime
correction, whereas MBPP (5.87\%), DROP (7.75\%), and
HotpotQA (2.88\%) usually retain the initial route. MATH
(98.97\%) and GSM8K (100\%) show the opposite pattern:
an early composite answer often makes the remaining suffix
unnecessary, so most edits reflect terminal pruning rather
than failure recovery. Edit rate therefore represents
corrective adaptation on HumanEval, early completion on
mathematics, and route stability on MBPP and QA.

\subsection{Ablation Study}

We conduct ablation on HumanEval with Qwen3-8B.
Each variant removes one source of offline structure or online
adaptation while leaving the remaining controller unchanged.
Relative to Full TROVE, \emph{Without Offline} removes both
composite skills and the transition graph and falls by 14.51
accuracy points while adding 9.26 minutes, showing the cost
of relying on atomic skills and online planning alone.
\emph{Without Composite} retains the graph but removes
composite execution; its similar loss of 13.74 points and
8.94-minute increase suggest that stable composite structure
accounts for most of the offline benefit in this setting.
\emph{Without Insert} retains suffix replacement but removes
local graph-supported repair, reducing accuracy by 9.93 points
and adding 14.74 minutes as intermediate failures require
broader replanning. \emph{Without Replace} retains insertion
but removes LLM suffix replanning; accuracy decreases by only
1.53 points, while time rises by 15.79 minutes. Thus,
replacement appears less critical for correctness than for
pruning or escaping unsuitable continuations efficiently.

\begin{table}[t]
  \centering
  \caption{Qwen3-8B HumanEval test ablation. Time is
  online test elapsed wall time.}
  \label{tab:ablation}
  \resizebox{0.9\linewidth}{!}{
  \begin{tabular}{lrr}
    \toprule
    Variant & Accuracy (\%) $\uparrow$ & Time (min) $\downarrow$ \\
    \midrule
    Full TROVE        & 92.37 & 24.36 \\
    Without Offline   & 77.86 & 33.62 \\
    Without Composite & 78.63 & 33.30 \\
    Without Insert    & 82.44 & 39.10 \\
    Without Replace   & 90.84 & 40.15 \\
    \bottomrule
  \end{tabular}
  }
\end{table}

\FloatBarrier

\section{Conclusion}

In this work, we have formulated continuation invalidation as a distinct orchestration problem and identified adaptive validation and minimal editing of an unexecuted route as the corresponding test-time requirement. We further proposed a trace-grounded framework TROVE that distills evaluated workflow-search trajectories into executable composite skills and an outcome-conditioned transition graph, and combines them with a retain--insert--replace controller that preserves completed work. Experiments across six benchmarks and three LLM backbones show that selectively revising a provisional skill route improves both task quality and online efficiency over dataset-level optimization, query-level pre-execution adaptation, and graph-constrained online scheduling. More broadly, TROVE advances agent orchestration from static workflow execution toward adaptive, experience-grounded control that efficiently responds to runtime feedback while preserving useful progress.

\begingroup
\raggedbottom
\setlength{\bibsep}{0pt plus 0.2ex}
\bibliography{references}
\endgroup

\end{document}